\documentclass[10pt,twocolumn,letterpaper]{article}

\usepackage{iccv}
\usepackage{times}
\usepackage{epsfig}
\usepackage{graphicx}
\usepackage{amsmath}
\usepackage{amssymb}
\usepackage{booktabs}
\usepackage {colortbl,array,xcolor}
\usepackage {url}

\usepackage[whole]{bxcjkjatype}
\usepackage{bm}
\usepackage[breaklinks=true,bookmarks=false]{hyperref}

\iccvfinalcopy % *** Uncomment this line for the final submission

\def\iccvPaperID{11} % *** Enter the ICCV Paper ID here

\ificcvfinal\fi

\begin{document}

%%%%%%%%% TITLE
\title{Defense-Prefix for Preventing Typographic Attacks on CLIP}

\author{Hiroki Azuma \qquad Yusuke Matsui\\
The University of Tokyo, Japan\\
{\tt\small $\lbrace$azuma, matsui$\rbrace$@hal.t.u-tokyo.ac.jp}
% \and
% Yusuke Matsui\\
% The University of Tokyo\\
% Tokyo, Japan\\
% {\tt\small matsui@hal.t.u-tokyo.ac.jp}
}

\maketitle
\ificcvfinal\thispagestyle{empty}\fi

\newcommand{\Fref}[1]{Fig.~\ref{#1}}
%%%% Added by Matsui %%%%%
\newcommand{\matsui}[1]{\textbf{\textcolor{cyan}{[\textsc{MATSUI:} #1]}}}
\newcommand{\azuma}[1]{\textbf{\textcolor{magenta}{[\textsc{AZUMA:} #1]}}}

%%%%%%%%% ABSTRACT
\begin{abstract}
   Vision-language pre-training models (VLPs) have exhibited revolutionary improvements in various vision-language tasks. In VLP, some adversarial attacks fool a model into false or absurd classifications. Previous studies addressed these attacks by fine-tuning the model or changing its architecture. However, these methods risk losing the original model's performance and are difficult to apply to downstream tasks. In particular, their applicability to other tasks has not been considered. In this study, we addressed the reduction of the impact of typographic attacks on CLIP without changing the model parameters. To achieve this, we expand the idea of ``class-prefix learning'' and introduce our simple yet effective method: Defense-Prefix (DP), which inserts the DP token before a class name to make words ``robust'' against typographic attacks. 
   Our method can be easily applied to downstream tasks, such as object detection, because the proposed method is independent of the model parameters. Our method significantly improves the accuracy of classification tasks for typographic attack datasets, while maintaining the zero-shot capabilities of the model. 
   In addition, we leverage our proposed method for object detection, demonstrating its high applicability and effectiveness. The codes and datasets are available at \url{https://github.com/azuma164/Defense-Prefix}.
   
\end{abstract}

\section{Introduction}
In recent years, vision-language pre-training models (VLPs) such as CLIP~\cite{Radford2021-bi} and ALIGN~\cite{jia2021align} have revolutionized downstream vision-language tasks such as classification~\cite{Conde2021art, zhang2022tip, Gao2021adapter}, object detection~\cite{zhong2022regionclip, feng2022promptdet}, segmentation~\cite{Zhou2022-tm, zhou2022maskclip}, and image generation~\cite{Ramesh2022-aw, Saharia2022-ma, crowson2022vqgan}. Such models are trained on web-scale data, for example, 400 million text-image pairs in the case of CLIP. The rich supervision provided by natural language enabled these pre-trained models to achieve impressive results on various downstream tasks with little or no additional training data.

\begin{figure}[t]
\begin{center}
   \includegraphics[width=1.0\linewidth]{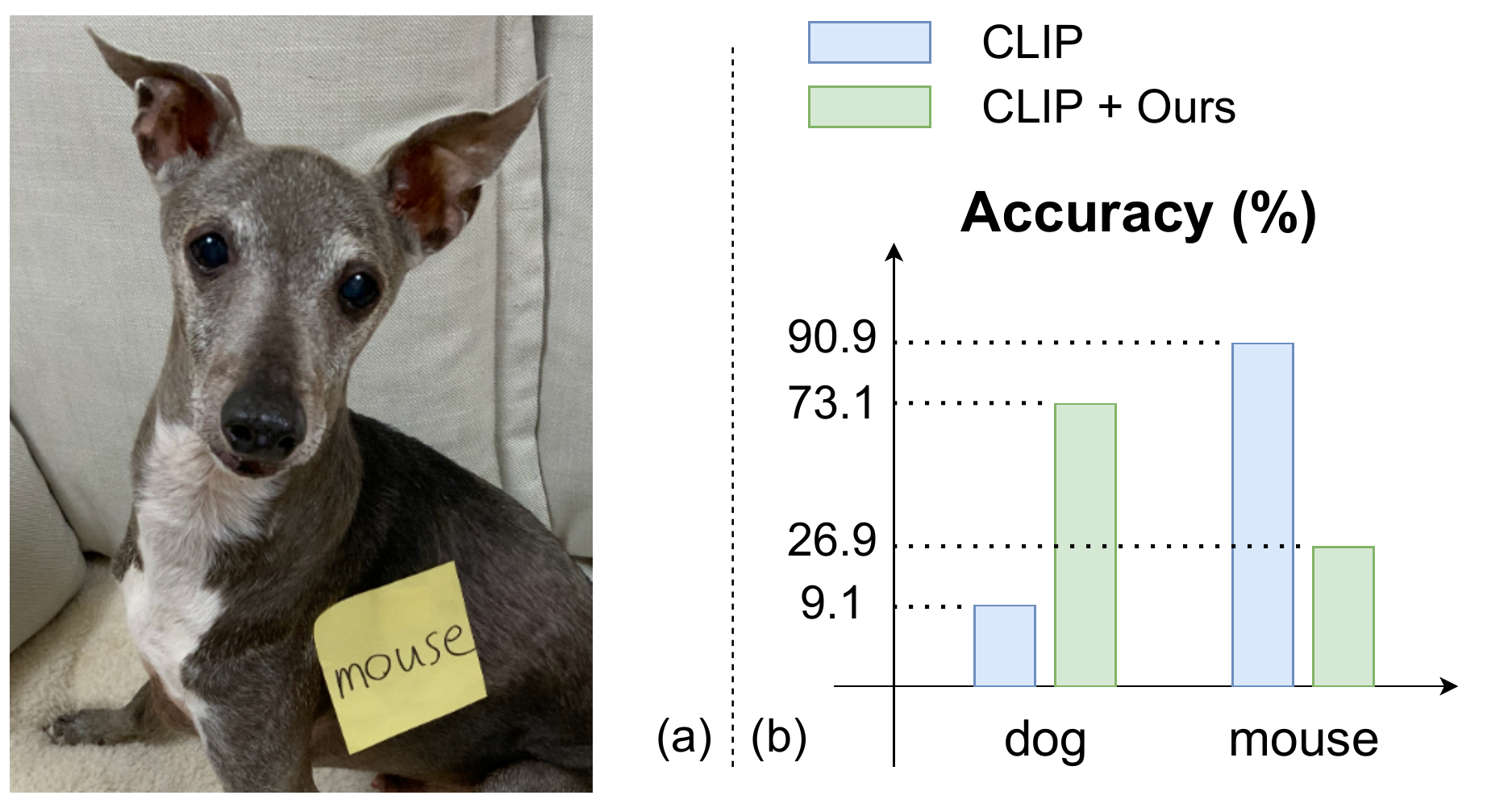}
\end{center}
   \caption{\textbf{(a): Image of a dog with a yellow tag that states ``mouse''. (b): Misclassification in CLIP against the image.}}
\label{fig:typographic_attacks}
\end{figure}

However, some adversarial attacks~\cite{jia2022badencoder, Goh2021-bn} can fool such models into making false or absurd classifications. Goh et al.~\cite{Goh2021-bn} found that CLIP is vulnerable to typographic attacks, in which the text in an image results in misclassification. In \Fref{fig:typographic_attacks}, the yellow tag that states ``mouse'' causes CLIP to misclassify the dog as a mouse. 

% To prevent typographic attacks, previous research has fine-tuned the models~\cite{Ilharco2022} or learned a transformation module on top of CLIP's output~\cite{Materzynska2022-pf}. However, these approaches risk forgetting prior knowledge or are hard to apply to downstream tasks. As described below, we find downstream classifiers built based on CLIP for different tasks such as object detection are also susceptible to typographic attacks, even if the image encoder is fine-tuned. Therefore, defense methods against the attacks should be readily applied to other downstream tasks. However, previous work has mainly focused on typographic attacks in classification and ignored this applicability. In addition, these approaches have to update the image features of CLIP. This can be a problem for image retrieval.

As described below, we found that downstream classifiers built based on CLIP for different tasks are also susceptible to typographic attacks. Therefore, defense methods against such attacks should be readily applied to other downstream tasks. However, previous studies~\cite{Ilharco2022, Materzynska2022-pf} have mainly focused on typographic attacks on classification and ignored their applicability. Materzynska et al.~\cite{Materzynska2022-pf} learned a transformation module on top of the CLIP output and PAINT~\cite{Ilharco2022} fine-tuned the model. Since these methods change the model parameters, they risk losing the original model's performance and are difficult to apply to downstream tasks. Additionally, if you calculate the image features of CLIP beforehand, these approaches require updating those features.
% This can also be a problem for retrieval tasks.

To solve these problems, we propose a simple yet effective defense method: Defense-Prefix (DP), which inserts the DP token before a class name. The DP token is a unique token followed by a class name (e.g., ``a photo of a [DP] dog''). 
An image feature from \Fref{fig:typographic_attacks}(a) would resemble a text feature from ``a photo of a mouse'', but would not be similar to a feature from ``a photo of a [DP] mouse''.
In other words, DP makes the class name ``robust'' against the attacks. Learning a unique token followed by a class name has been primarily conducted in subject-driven image generation~\cite{Ruiz2022-oy, kumari2022customdiffusion, Lin2022magic3d}. We define this approach as \textit{class-prefix learning} and apply the concept of class-prefix learning to prevent typographic attacks.

Our approach learns only the word embedding vector for the DP token. Therefore, we do not update the original CLIP. After the DP vector is obtained, it can be used for any task. This simplicity is a significant advantage over existing works because all other works require training the model. 
% In addition, we can easily apply our method to other downstream tasks without additional training, since the method is independent of the model of the downstream tasks.

We experimentally demonstrate the effectiveness of the proposed method. \textbf{(1)} We first conduct experiments on classification using ten synthetic and three real-world typographic attack datasets. Here, due to the insufficient number of datasets, we create the biggest \underline{R}eal-world \underline{T}ypographic \underline{A}ttack dataset ``RTA-100'', which contains 100 categories and 1000 images. Compared with CLIP, our method effectively prevents typographic attacks (e.g., +9.61\% on synthetic and +17.70\% on real-world datasets), while losing only 0.64\% on average for original datasets. \textbf{(2)} We also evaluate our method on object detection by using RegionCLIP~\cite{zhong2022regionclip}. The proposed method does not require additional training because only the input of the text encoder is modified. Our results indicate that the downstream classifiers based on CLIP are also susceptible to typographic attacks. Our method reduces the impact of the attacks (e.g., +16.0 AP50 on COCO, +6.2 mAP on LVIS), while keeping the original accuracy (e.g., +0.1 AP50 on COCO, -0.3 mAP on LVIS).

In summary:
\begin{itemize}
\item We expand class-prefix learning and propose DP, a novel method for preventing typographic attacks on CLIP without changing the model parameters.
\item We find downstream classifiers built based on CLIP are also vulnerable to typographic attacks.
\item Our method effectively prevents typographic attacks, while keeping the original model's performance. In addition, we demonstrate the easy application of our approach to downstream tasks. 
\item We creat the biggest real-world typographic attack dataset RTA-100, which will be publicly available.
\end{itemize}

\section{Related work}
\subsection{Vision-language pre-training (VLP)}
Learning the joint vision-language representation space has been of great interest in the field of computer vision. Recently, CLIP~\cite{Radford2021-bi} and ALIGN~\cite{jia2021align} collected million/billion-scale image-caption pairs from the Internet and learned to match images with image descriptions. These models obtain a strong vision-language representation space, which has been extremely effective for downstream tasks.

Recent studies have transferred the knowledge of these models to downstream recognition tasks, such as classification~\cite{Conde2021art, zhang2022tip, Gao2021adapter}, object detection~\cite{zhong2022regionclip, feng2022promptdet}, semantic segmentation~\cite{zhou2022maskclip, Zhou2022-tm}, panoptic segmentation~\cite{Ding2022}, and multi-label recognition~\cite{Sun2022dualcoop}. 
Typically, these methods freeze a VLP text encoder and then use it directly. Therefore, the proposed method can be applied without additional training.

\subsection{Typographic attacks}
CLIP is known to be weak against typographic attacks~\cite{Goh2021-bn, Avrahami2022-ta-image-generation}. Goh et al.~\cite{Goh2021-bn} found that the text in an image results in misclassification of CLIP as shown in \Fref{fig:typographic_attacks}.

Materzynska et al.~\cite{Materzynska2022-pf} applied the learned linear transformation to the CLIP output to disentangle the visual concept from the spelling capabilities of CLIP. Ilhalco et al.~\cite{Ilharco2022} interpolated the weights of the parameters between the fine-tuned and the original CLIP models to prevent typographic attacks. These methods risk losing the original model's performance and are difficult to apply to downstream tasks. Also, they need to update the image features.

Unlike these methods, our method does not modify the architecture or model parameters. In addition, our method does not update the image features.

\subsection{Prompt learning in VLP}
% To adapt VLP to downstream tasks, two typical approaches are fine-tuning the model~\cite{Conde2021art} or learning the transformation module to VLP's output~\cite{zhang2022tip, Gao2021adapter}. However, these methods risk losing the original model's performance. 

Inspired by the success in NLP~\cite{autoprompt:emnlp20, jiang2020prompt, zhong2021factual}, to adapt VLP to downstream tasks, several studies have learned prompt tokens in end-to-end training.
% ~\cite{Zhou2022-ai, Zhou2022-ew, shu2022tpt, Sun2022dualcoop, Zhou2022-tm, feng2022promptdet, zhou2022maskclip} 
CoOp~\cite{Zhou2022-ai} first utilized prompt learning in VLP to improve the accuracy of classification tasks. This was followed by other studies~\cite{Zhou2022-ew, shu2022tpt, Khattak2022maple}. Recently, some studies~\cite{Sun2022dualcoop, Zhou2022-tm, feng2022promptdet, zhou2022maskclip, du2022detpro} have focused on using prompt learning to improve other downstream recognition tasks apart from classification.

Prompt learning trains tokens of the whole sentence except for a class name, whereas our class-prefix learning trains one token before a class name. Tokens obtained by class-prefix learning can be used for any task that uses prompts to input text, whereas prompt learning must be trained only for the specific recognition task and cannot be used for any other task. 

\subsection{Class-prefix learning}
We define the approach for learning a unique token followed by a class name as \textit{class-prefix learning}. 
% As aforementioned, prefix learning trains only one token, whereas prompt learning trains all tokens except class names.  
Class-prefix learning has been mainly conducted in the research of image generation~\cite{Ruiz2022-oy, kumari2022customdiffusion, Lin2022magic3d, schwartz2023discriminative}. Ruiz et al.~\cite{Ruiz2022-oy} addressed a new problem: subject-driven generation. They learned a unique identifier followed by the class name of the subject (e.g., ``A [V] dog''). They aimed to synthesize novel scenes of the subject in different contexts while keeping its key visual features.

Apart from image generation, class-prefix learning has rarely been investigated. Because class-prefix learning retains the original input texts, it can be incorporated into various vision-language tasks. In this study, we propose a novel method for learning a prefix to prevent typographic attacks.

\begin{figure*}[t]
\begin{center}
    \includegraphics[width=0.85\linewidth]{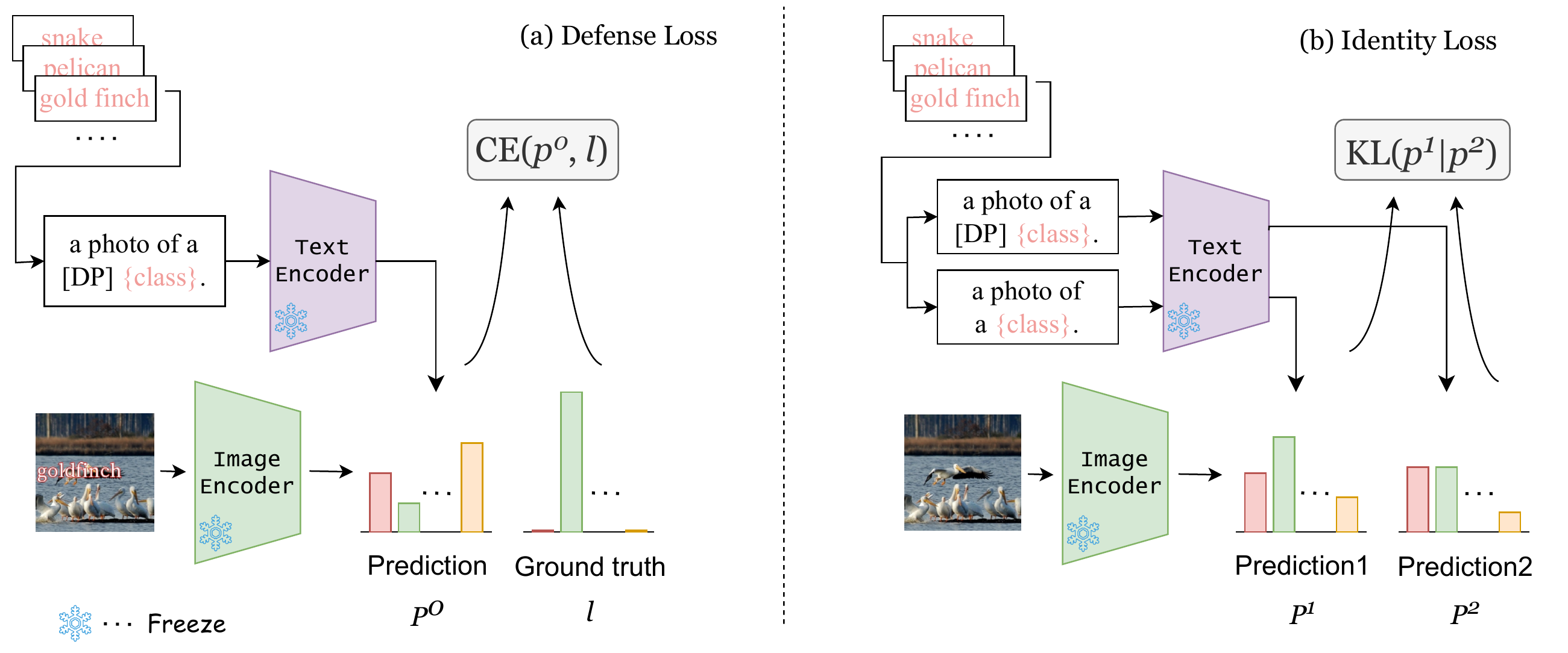}
\end{center}
    \caption{\textbf{Method overview.} We keep the image encoder and text encoder of CLIP frozen. Our method trains only the DP vector, which is a word embedding for [DP]. We propose to learn the DP vector by using \textit{Defense loss} and \textit{Identity loss}. \textbf{(a)} \textit{Defense loss} calculates cross-entropy loss against typographic attack images. \textbf{(b)} \textit{Identity loss} calculates KL-divergence loss between two probability distributions.}
\label{fig:proposed_method}
\end{figure*}

\section{Method}
\subsection{Preliminaries: CLIP}
% ref coop, dreambooth
We first introduce CLIP~\cite{Radford2021-bi} as the basis for our approach. It consists of two encoders: an image encoder and a text encoder. CLIP encodes the images and text in the same embedding space. The image encoder can be either ResNet~\cite{He2016-zx} or Vision-Transformer~\cite{Dosovitskiy2020-vz}. The text encoder is Transformer~\cite{Vaswani2017-kq}. To encode an input text, such as ``a photo of a dog'', CLIP first converts each word to a $d$-dimensional word embedding vector ($d$ represents the dimension of a word embedding vector), using a learned vocabulary. Subsequently, the word embedding vectors are fed into the transformer to obtain the final text feature.

The CLIP can be used for zero-shot image recognition. Let us consider $n$-class image recognition problem. Let $\mathbf{x}\in\mathbb{R}^{m}$ be an image feature generated by the image encoder ($m$ represents the dimension of a feature vector) and $\{\mathbf{w}_i\}_{i=1}^n$ be a set of text features produced by the text encoder. Here, $\mathbf{w}_i \in \mathbb{R}^{m}$ represents the $i$-th category.
In particular, each $\mathbf{w}_i$ is derived from a text prompt based on a template such as ``a photo of a \verb|<CLS>|.'', where \verb|<CLS>| can be replaced with the $i$-th class name. The prediction probability that the output label $y$ is of class $i$ is then
\begin{equation}
    p(y = i \mid \mathbf{x}, \{\mathbf{w}_j\}_{j=1}^n) = \frac{\exp{(\cos{(\mathbf{w}_i, \mathbf{x})} / \tau)}}{\sum_{j=1}^{n}\exp{(\cos{(\mathbf{w}_j, \mathbf{x})} / \tau)}},
\label{eq:clip-recognition}
\end{equation}
where $\cos{(\cdot,\cdot)}$ calculates the cosine similarity and $\tau$ is a temperature parameter learned by CLIP.

\subsection{Defense-Prefix}
\label{defense-prefix}
In this section, we present the proposed approach. Our goal is to train the word embedding vector for the DP token, i.e., a single $d$-dimensional vector. We define this word embedding vector as the DP vector. Here, none of the model parameters are modified. 
Given the $i$-th class name, we define the input sequence of words (text prompts) as $t_i$. We also prepare $t^\mathrm{DP}_i$, which contains the DP token. 
\begin{eqnarray}
    t_i &=& \left(\mathrm{P}_1, \mathrm{P}_2, ..., \mathrm{CLS}_i, ..., \mathrm{P}_l\right). \\
    t_i^\mathrm{DP} &=& \left(\mathrm{P}_1, \mathrm{P}_2, ..., \left[DP\right], \mathrm{CLS}_i, ...,  \mathrm{P}_l\right).
\end{eqnarray}
Here, $\left[DP\right]$ and $\mathrm{CLS}_i$ represent the DP token and $i$-th class name, respectively, while $\mathrm{P}_1, \mathrm{P}_2, \dots$ form a template of $l$ words. For example, in the case ``a photo of a \verb|<CLS>|.'', $\mathrm{P}_1$ is ``a'' and $\mathrm{P}_2$ is ``photo''. As aforementioned, CLIP converts each word into a $d$-dimensional word embedding vector using the learned vocabulary as follows:
\begin{eqnarray}
    b_i &=& \left(\mathbf{B}_{\mathrm{P}_1}, \mathbf{B}_{\mathrm{P}_2}, ..., \mathbf{B}_{\mathrm{CLS}_i}, ..., \mathbf{B}_{\mathrm{P}_l}\right). \\
    b_i^\mathrm{DP} &=& \left(\mathbf{B}_{\mathrm{P}_1}, \mathbf{B}_{\mathrm{P}_2}, ..., \mathbf{B}_{\left[DP\right]}, \mathbf{B}_{\mathrm{CLS}_i}, ...,  \mathbf{B}_{\mathrm{P}_l}\right),
\end{eqnarray}
where $\mathbf{B}_{\mathrm{P}_1}, \mathbf{B}_{\mathrm{P}_2}, \dots, \mathbf{B}_{\mathrm{CLS}_i} \in \mathbb{R}^{d}$ denote the learned word embedding vectors. The vectors are pre-trained and fixed.
Here, we aim to learn the DP vector ($\mathbf{B}_{\left[DP\right]} \in \mathbb{R}^{d}$), which is a word embedding vector for the DP token.

Then, we enter $\{b_i\}_{i=1}^n$ and $\{b_i^\mathrm{DP}\}_{i=1}^n$ into the text encoder and obtain the original and ``robust'' class features $\{\mathbf{w}_i\}_{i=1}^n$ and $\{\mathbf{w}^\mathrm{DP}_i\}_{i=1}^n$, respectively. Here, $n$ represents the number of classes and all $\mathbf{w}_i, \mathbf{w}_i^\mathrm{DP} \in \mathbb{R}^{m}$. We can now recognize an image using Eq.~\ref{eq:clip-recognition} with the original ($\{\mathbf{w}_i\}_{i=1}^n$) or the robust ($\{ \mathbf{w}_i^\mathrm{DP} \}_{i=1}^n$) class features. Robust class features reduce the impact of typographic attacks.

The goal is to train the DP vector so that the word next to the DP token is robust against typographic attacks. To achieve this, we propose using \textit{defense loss} and \textit{identity loss} (\Fref{fig:proposed_method}). Defense loss enables the DP token to prevent typographic attacks, and identity loss helps it maintain the original meanings of the class names. For the training, we assume that a set of image pairs, comprising original and ``attack'' images, is available. The attack image is obtained by synthesizing the incorrect label text on the original image. We calculate defense loss and identity loss for each pair.
\paragraph{Defense loss:}
The defense loss aims to prevent typographic attacks. To achieve this, we adopt the cross-entropy loss in the same manner as for ordinary classification tasks. Let $I$ and $\bar{I}$ represent the original and attack images, respectively. For example, $I$ and $\bar{I}$ show an image of a dog and the same image of the same dog but with a synthesized text ``bird'', respectively. 
We then obtain the image feature $\bar{\mathbf{x}}$ by applying $\bar{I}$ to the image encoder. We classify the typographic attack image $\bar{I}$ using robust class features $\{\mathbf{w}_i^\mathrm{DP}\}_{i=1}^n$ as follows:
\begin{equation}
    p^0(y = i \mid \bar{\mathbf{x}}, \{\mathbf{w}_j^{\mathrm{DP}}\}_{j=1}^n)) = \frac{\exp{(\cos{(\mathbf{w}_i^{\mathrm{DP}}, \bar{\mathbf{x}}) / \tau)}}}{\sum_{j=1}^{n}\exp{(\cos{(\mathbf{w}_j^{\mathrm{DP}}, \bar{\mathbf{x}})} / \tau)}}.
\end{equation}
We minimize the standard classification loss based on the cross-entropy to train the DP vector. 
The defense loss for $\bar{I}$ is computed as follows:
\begin{eqnarray}
    % L_0 = - \sum_{j=1}^{J} y_k \log p^0(y=t) = -\log(p^0(y=t\mid \bar{\mathbf{x}}, \{\mathbf{w}_j^{\mathrm{DP}}\}_{j=1}^n))
    % L_0 = - \sum_{j=1}^{n} l_j \log p^0(y=j) = -\log(p^0(y=t)),
    L_0 = - \sum_{j=1}^{n} l_j \log p^0(y=j),
\end{eqnarray}
where $\bm{l}$ is a one-hot vector representing the ground truth.

\paragraph{Identity loss:}
The identity loss function aims to help the learned token maintain the original meanings of the words. To achieve this goal, we ensure a consistent output with and without DP tokens. To distill the knowledge of CLIP, some studies~\cite{Gu2022vild, Ma2022-distill} have used the output features of CLIP. However, how to use text features for distillation in our method is unclear. Then, we utilize classification results. First, we classify the original image $I$ using the original ($\{\mathbf{w}_i\}_{i=1}^n$) and robust ($\{\mathbf{w}_i^\mathrm{DP}\}_{i=1}^n$) class features as follows:
\begin{equation}
    p^1(y = i \mid \mathbf{x}, \{\mathbf{w}_j\}_{j=1}^n) = \frac{\exp{(\cos{(\mathbf{w}_i, \mathbf{x})}  
    / \tau)}}{\sum_{j=1}^{n}\exp{(\cos{(\mathbf{w}_j, \mathbf{x})} / \tau)}}.
\label{eq:original}
\end{equation}
\begin{equation}
    p^2(y = i \mid \mathbf{x}, \{\mathbf{w}_j^{\mathrm{DP}}\}_{j=1}^n) = \frac{\exp{(\cos{(\mathbf{w}_i^{\mathrm{DP}}, \mathbf{x})} / \tau)}}{\sum_{j=1}^{n}\exp{(\cos{(\mathbf{w}_j^{\mathrm{DP}}, \mathbf{x})} / \tau)}},
\label{eq:recognition}
\end{equation}
where $\mathbf{x}$ denotes the image feature from $I$. Here, we make the probability distribution of $\{p^2\}_{i=1}^n$ approach that of $\{p^1\}_{i=1}^n$ using KL-divergence. Formally, the identity loss for $I$ is defined as:
\begin{equation}
    L_1=D_\mathrm{KL}\left[\sum_{j=1}^{n}p^1(y=j) \mathbf{e}_j \parallel \sum_{j=1}^{n}p^2(y=j) \mathbf{e}_j\right],
\end{equation}
where $\mathbf{e}_j$ is a one-hot vector ($j$-th element is one). DP maintains the performance of the original model by mimicking the original classification results.

Finally, the loss for the image pair $\{I, \bar{I}\}$ is computed as:
\begin{equation}
    L = L_0 + \lambda L_1,
\end{equation}
where $\lambda$ is a hyperparameter that balances the losses. Empirically, we set $\lambda=3.0$.

It is worth noting that our method does not modify any parameters of the image and text encoders of CLIP but trains only the DP vector. Originally, CLIP recognizes images using Eq.~\ref{eq:original}. In our method, after training the DP vector, we use it to apply various recognition tasks using Eq.~\ref{eq:recognition}.

\section{Experiments}
\subsection{Training Defense-Prefix}
\label{section:training}
First, we train the DP vector. After obtaining the learned DP vector, we apply it to the experiments of recognition tasks in Sec.~\ref{section:classification} and \ref{section:object-detection}. We train the DP vector only in Sec.~\ref{section:training}.
\paragraph{Datasets:}
We use ImageNet-100~\cite{ImageNet-100}, a random 100-class subset of ImageNet~\cite{Deng2009-rk}, to train the DP vector. We generate typographic attack images by adding text with incorrect labels to the original images.
\paragraph{Implementation details:}
We initialize the image and text encoders from the CLIP~\cite{Radford2021-bi} pre-trained model and keep them frozen during training. For the image encoder, ViT-B/32 and RN50x4 are applied for classification and object detection, respectively. We train only one vector for DP, which is the only learnable part of our method.
The DP vector is randomly initialized by drawing from a zero-mean Gaussian distribution with a standard deviation of 0.02. We use SGD optimizer with an initial learning rate of 0.002, which is decayed using the cosine annealing rule. We train the DP vector for 10 epochs with a batch size of 512, using one NVIDIA V100.

\subsection{Classification}
\label{section:classification}
\begin{figure}[t]
\begin{center}
   \includegraphics[width=1.0\linewidth]{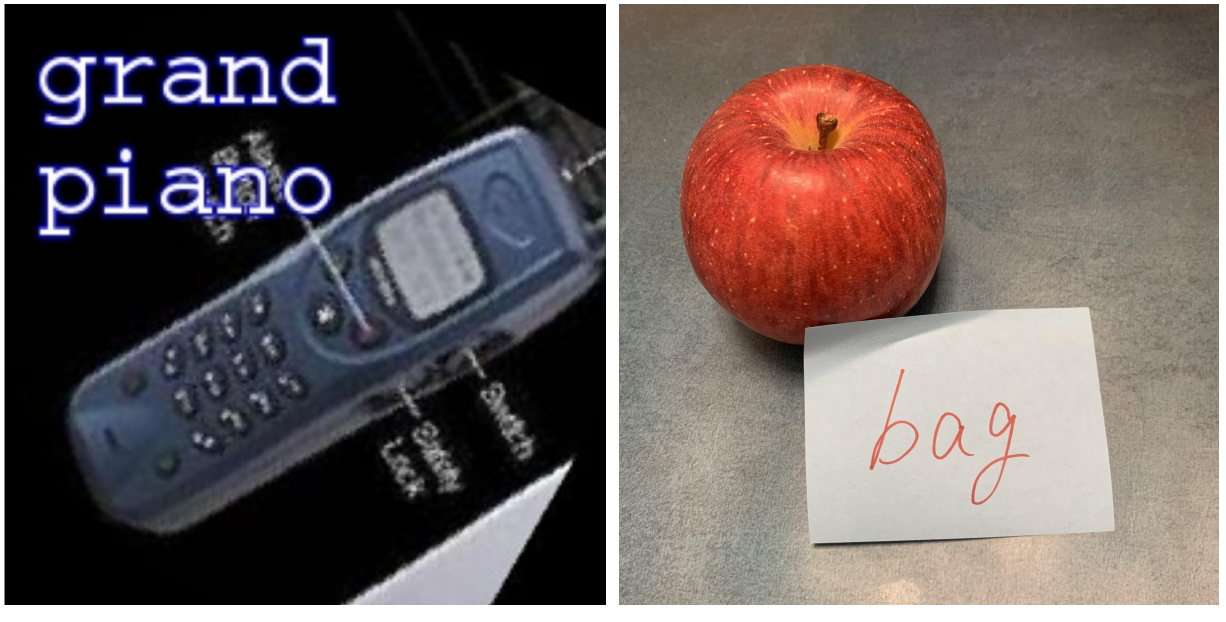}
\end{center}
   \caption{\textbf{Typographic attack datasets.} (Left: a sample from synthetic typographic attack datasets, Right: a sample from our real-world typographic attack dataset.)}
   \label{fig:datasets}
\end{figure}

In this section, we evaluate the performance of the proposed method based on the classification tasks. We compare our method to CLIP~\cite{Radford2021-bi}, Materzynska et al.~\cite{Materzynska2022-pf}, and PAINT~\cite{Ilharco2022}. 
\paragraph{Datasets:}
We employ ten publicly available image classification datasets used in CLIP: ImageNet~\cite{Deng2009-rk}, Caltech101~\cite{Fei2004caltech}, OxfordPets~\cite{Parkhi2012pets}, StanfordCars~\cite{Krause2013cars}, Flowers102~\cite{Nilsback2008flowers}, Food101~\cite{Bossard2014food}, FGVCAircraft~\cite{Maji2013aircraft}, DTD~\cite{Cimpoi2014dtd}, SUN397~\cite{xial2018sun}, EuroSAT~\cite{helber2019eurosat}. 
To evaluate the classification of typographic attack datasets, we create synthetic typographic attack datasets using those ten datasets (\Fref{fig:datasets}: left). Also, we use two publicly available real-world typographic attack datasets from Materzynska et al.~\cite{Materzynska2022-pf} and PAINT. In addition, due to the insufficient number of datasets, we generate our real-world attack dataset RTA-100 (\Fref{fig:datasets}: right). For real-world attack datasets, we use class labels of objects and labels of tags as the candidate categories.

\paragraph{RTA-100:}
As described before, we create the biggest real-world typographic attack dataset RTA-100, which contains 100 categories and 1000 images. The dataset from Materzynska et al.~\cite{Materzynska2022-pf} comprises 19 categories and 171 images, and that from PAINT~\cite{Ilharco2022} has 89 categories and 110 images. Combining those datasets is not sufficient to verify the diversity. To increase the test data, we created RTA-100 (see Appendix for more details).

\paragraph{Implementation details:}
We use ViT-B/32 for the image encoder. When we evaluate our method on classification, we place the DP token before the class names. 

\paragraph{Baselines:}
To evaluate the effectiveness of the proposed method, we compare it with the following baselines: CLIP~\cite{Radford2021-bi}, Materzynska et al.~\cite{Materzynska2022-pf}, and PAINT~\cite{Ilharco2022}. Materzynska et al.~\cite{Materzynska2022-pf} apply the learned linear layer to the CLIP output. For Materzynska et al.~\cite{Materzynska2022-pf}, we use a publicly available pre-trained linear layer for ViT-B/32. This linear layer was trained using ImageNet-1K and 182,329 English words. We apply the linear layer to the output of both the image and text encoders of CLIP. For PAINT, we fine-tune the image encoder of CLIP using typographic attack images from ImageNet-100, which is used to train the DP vector. 
We then interpolate the weights between the fine-tuned image encoder $\theta_{ft}$ and the original image encoder $\theta_{zs}$ with $\alpha=0.35$, where $\alpha$ is the mixing coefficient ($\alpha\in[0,1]$). We get \textit{patched model} as follows:
$\theta_{patch}=(1-\alpha)\theta_{zs}+\alpha\theta_{ft}$.

\begin{table}[t]
\begin{center}
\small
\caption{\textbf{Summary of classification results.} The best results out of Materzynska +, PAINT, and ours are \textbf{bolded}.}
\label{table:all}
\begin{tabular}{@{}llllll@{}} \toprule
     & Retain & & \multicolumn{3}{c}{Typographic attack} \\ \cmidrule(l) {4-6} 
    Method & Models & Original & Synth. & Real & Avg. \\ \midrule
    CLIP & - & 61.55 & 34.59 & 46.82 & 40.71\\ \midrule
    Materzynska+~\cite{Materzynska2022-pf} & $\times$ & 49.50 & 37.44 & 63.61 & 50.53 \\
    PAINT~\cite{Ilharco2022} & $\times$ & 59.63 & \textbf{49.93} & 55.00 & 52.47\\ 
    Ours & $\checkmark$ & \textbf{60.91} & 44.20 & \textbf{64.52} & \textbf{54.36}\\ \bottomrule
\end{tabular}
\end{center}
\end{table}

\begin{table*}[t]
\begin{center}
% \small
\caption{\textbf{Classification results on original datasets.} Individual results for all 10 datasets are available in the Appendix. $^*$Average reported across 10 datasets.}
\label{table:original}
\begin{tabular}{@{}llllllll@{}} \toprule
    Method & Retain models & ImageNet & Caltech & Pets & Cars & $^*$Avg. \\ \midrule
    CLIP & - & 62.02& 88.64& 87.35& 58.72& 61.55 \\ \midrule
    Materzynska+~\cite{Materzynska2022-pf} & $\times$& 54.38& 80.53& 75.01& 40.33& 49.50 \\
    PAINT~\cite{Ilharco2022} & $\times$& 61.82& 88.48& 85.23& 55.30& 59.63\\ 
    Ours & $\checkmark$& \textbf{62.48}& \textbf{89.28}& \textbf{87.22}& \textbf{57.47}& \textbf{60.91}\\ \bottomrule
\end{tabular}
\end{center}

\begin{center}
\caption{\textbf{Classification results on typographic attack datasets.} $^*$Average reported across 10 datasets.}
\small
\label{table:realta}
\begin{tabular}{@{}lllllllllll@{}} \toprule
     & & \multicolumn{5}{c}{Synth.} & \multicolumn{4}{c}{Real} \\ \cmidrule(l) {3-7} \cmidrule(l) {8-11}
    Method & Retain models & ImageNet & Caltech & Pets & Cars & $^*$Avg.& from ~\cite{Materzynska2022-pf} & from ~\cite{Ilharco2022} & RTA-100 & Avg. \\ \midrule
    CLIP & - & 39.10& 63.97& 58.95& 21.02 & 34.59& 43.27 & 50.00 & 47.20 & 46.82\\ \midrule
    Materzynska+~\cite{Materzynska2022-pf} & $\times$ & 44.91& 74.73& 63.61& 15.79& 37.44& \textbf{77.78} & 55.45 & 57.60 & 63.61\\
    PAINT~\cite{Ilharco2022} & $\times$ & \textbf{55.9}& \textbf{83.57}& \textbf{76.53}& \textbf{33.44}& \textbf{49.93}& 53.22 & 58.18 & 53.60 & 55.00\\ 
    Ours & $\checkmark$ & 49.83& 79.54& 72.88& 28.64& 44.20& 71.93 &  \textbf{63.64} & \textbf{58.00} & \textbf{64.52}\\ \bottomrule
\end{tabular}
\end{center}
\end{table*}

\paragraph{Results:}
Table~\ref{table:all} summarizes the performance of our method on classification. 
As previous research~\cite{Goh2021-bn} has shown, our results demonstrate that text in images harms the original performance of CLIP (e.g., from 61.55\% to 34.59\% on average). Compared with CLIP, our method improves the performance on all typographic attack datasets (e.g., from 34.59\% to 44.20\% on synthetic and from 46.82\% to 64.52\% on real-world datasets), losing little average accuracy on the original datasets (e.g., from 61.55\% to 60.91\%).

Compared to Materzynska et al., our method exhibits improved performance on both synthetic and real-world typographic attack datasets (e.g., from 37.44\% to 44.20\% on synthetic and from 63.61\% to 64.52\% on real-world datasets). When compared with PAINT, our method loses on synthetic attack datasets (e.g., from 49.93\% to 44.20\% on average), while it significantly improves the performance on real-world attack datasets (e.g., from 55.00 to 64.52 on average). The result indicates that our method is more robust against changes in the appearance of text.

Tables~\ref{table:original} and ~\ref{table:realta} present the specific performance in classifying original datasets, and typographic attack datasets, respectively.

Overall, our simple method effectively prevents typographic attacks (e.g., +9.61\% on synthetic and +17.70\% on real-world typographic attack datasets), while losing the least original accuracy (e.g., -0.64\% on average). Although our method does not update CLIP, our simple method of putting the learned prefix before the class names works effectively, even when compared to previous studies. Here, it is worth noting that PAINT must retrain the CLIP encoder and recompute the CLIP features for all images to achieve typographic defense. In contrast, our approach does not need to modify the encoder or existing features. This property is a clear advantage; we can apply our method to any CLIP-based application without modification. Therefore, our method is much better than PAINT if the performance is comparable to PAINT.

\subsection{Object detection}
\label{section:object-detection}
\begin{figure*}[t]
\begin{center}
    \includegraphics[width=1.0\linewidth]{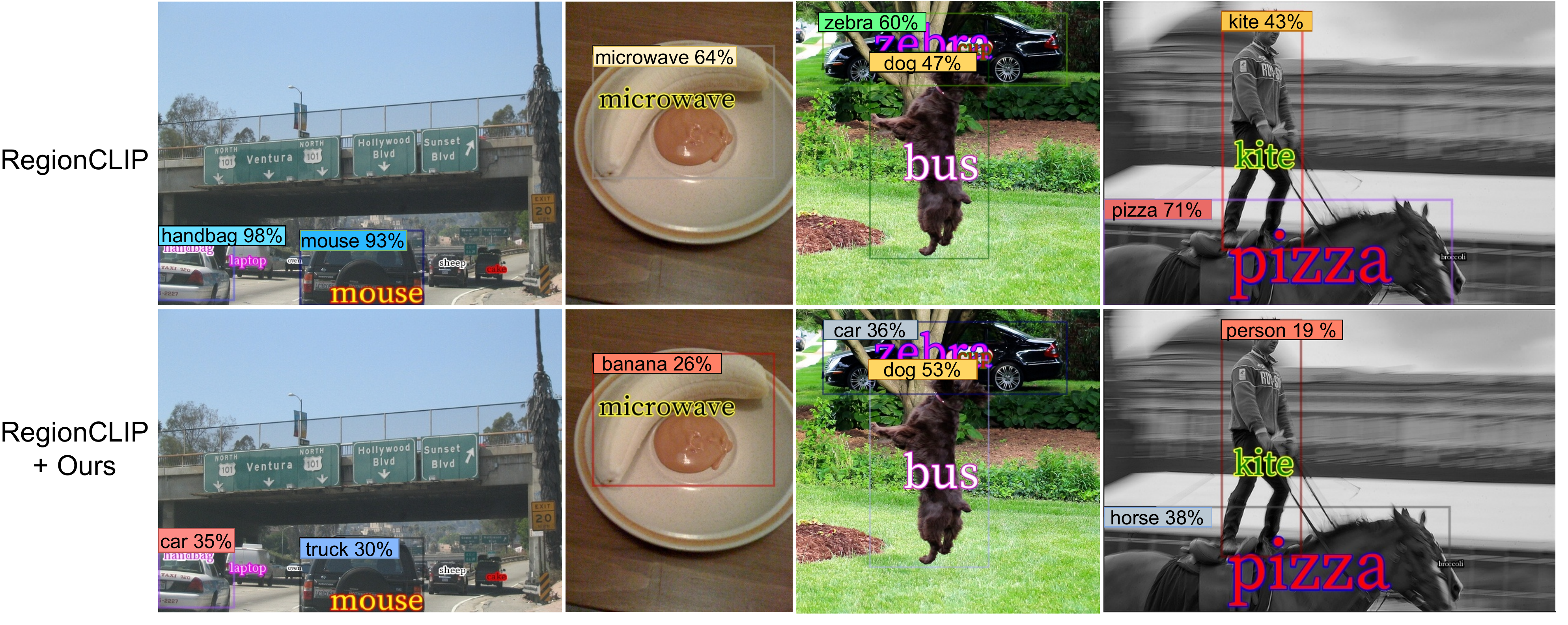}
\end{center}
    \caption{\textbf{Visualization of RegionCLIP and RegionCLIP+Ours zero-shot inference on the typographic attack COCO dataset with ground-truth boxes} (top: RegionCLIP, bottom: RegionCLIP+Ours). The pre-trained models are adversely affected by texts in images. Our proposed method reduces the impact of typographic attacks. (Image IDs: 1532, 13004, 17029, 23126)}
    \label{fig:misclassification_regionclip}
\end{figure*}

In this section, we evaluate the applicability of the proposed method to downstream tasks. In particular, we apply our method to RegionCLIP~\cite{zhong2022regionclip}, a zero-shot object detection model. In RegionCLIP, the image encoder is fine-tuned from the CLIP image encoder.
Therefore, we cannot apply previous methods~\cite{Materzynska2022-pf, Ilharco2022} directly to RegionCLIP because they need to update the model.
On the other hand, we can use DP directly, which we train in Sec.~\ref{section:training}, because it is independent of the parameters of the image encoder. 
\paragraph{Datasets:}
We evaluate our method through object detection experiments in COCO~\cite{lin2014coco} and LVIS~\cite{gupta2019lvis} for zero-shot inference. We use the standard object detection metrics (AP50 for COCO and mAP for LVIS). We create typographic attack datasets using COCO and LVIS by synthesizing text in each bounding box.

\paragraph{Implementation details:}
We use a pre-trained RegionCLIP model for RN50x4. We keep the model frozen during the inference and only modify the input of the text encoder by placing the DP token before the class names.

Following RegionCLIP, we evaluate two settings: (1) Ground-truth (GT) bounding boxes used as region proposals. (2) Region proposals obtained from RPN~\cite{NIPS2015-RPN}.
\paragraph{Baselines:}
We use RegionCLIP for zero-shot object detection. The model was pre-trained on Conceptual Caption dataset (CC3M)~\cite{sharma2018cc3m} using the concepts parsed from COCO Caption (COCO cap)~\cite{chen2015cococap}. 
RegionCLIP comprises an RPN and an image encoder. First, possible image regions are proposed by RPN. The model then calculates the similarity between the image features of the proposed regions and the text features of the target categories, recognizing the categories within the local image regions.

\paragraph{Results:}
% \begin{table}[t]
% \begin{center}
% \caption{Zero-shot object detection on original datasets}
% \small
% \label{table:regionclip-origin}
% \begin{tabular}{@{}llll@{}} \toprule
%      & Region & COCO & LVIS \\ 
%     Method & Proposals & All & mAP \\ \midrule
%     RegionCLIP & GT & \textbf{65.0} & \textbf{50.2}\\
%     RegionCLIP+Ours & GT & 64.4 & 49.4\\ \midrule
%     RegionCLIP & RPN & 14.5 & 11.1\\
%     RegionCLIP+Ours & RPN & \textbf{14.7} & \textbf{11.3}\\ \bottomrule
% \end{tabular}
% \end{center}

\begin{table}[t]
\begin{center}
\caption{\textbf{Zero-shot object detection on original datasets}}
\small
\label{table:regionclip-origin}
\begin{tabular}{@{}llll@{}} \toprule
     & Region & COCO & LVIS \\ 
    Method & Proposals & AP50 & mAP \\ \midrule
    RegionCLIP & GT & 65.5 & \textbf{50.2}\\
    RegionCLIP+Ours & GT & \textbf{65.6} & 49.9 \\ \midrule
    RegionCLIP & RPN & \textbf{29.6} & 11.1\\
    RegionCLIP+Ours & RPN & \textbf{29.6} & \textbf{11.3} \\ \bottomrule
\end{tabular}
\end{center}

% \begin{center}
% \caption{Zero-shot object detection on typographic attack datasets}
% \small
% \label{table:regionclip-ta}
% \begin{tabular}{@{}llll@{}} \toprule
%      & Region & COCO & LVIS \\
%     Method & Proposals & All & mAP \\ \midrule
%     RegionCLIP & GT & 24.5 & 31.9\\
%     RegionCLIP+Ours & GT & \textbf{40.6} & \textbf{37.8}\\ \midrule
%     RegionCLIP & RPN & 4.46 & 5.17\\
%     RegionCLIP+Ours & RPN & \textbf{6.04} & \textbf{6.21}\\ \bottomrule
% \end{tabular}
% \end{center}
% \end{table}

\begin{center}
\caption{\textbf{Zero-shot object detection on typographic attack datasets}}
\small
\label{table:regionclip-ta}
\begin{tabular}{@{}llll@{}} \toprule
     & Region & COCO & LVIS \\
    Method & Proposals & AP50 & mAP \\ \midrule
    RegionCLIP & GT & 25.0 & 31.9\\
    RegionCLIP+Ours & GT & \textbf{41.0} & \textbf{38.1} \\ \midrule
    RegionCLIP & RPN & 11.0 & 5.17\\
    RegionCLIP+Ours & RPN & \textbf{14.4} & \textbf{6.25} \\ \bottomrule
\end{tabular}
\end{center}
\end{table}

\Fref{fig:misclassification_regionclip} visualizes the results of zero-shot inference of RegionCLIP and RegionCLIP+Ours with GT boxes on the typographic attack COCO dataset. This shows RegionCLIP is also adversely influenced by typographic attacks, although the image encoder is fine-tuned. For example, the car is misclassified as a handbag (\Fref{fig:misclassification_regionclip}: top left). However, RegionCLIP+Ours correctly recognizes the car.

Tables~\ref{table:regionclip-origin} and ~\ref{table:regionclip-ta} present the performance of RegionCLIP and RegionCLIP+Ours. When using GT boxes, compared with the original RegionCLIP, our method shows improved performance on COCO and LVIS for the typographic attack datasets (e.g., 41.0 vs. 25.0 on COCO, 38.1 vs. 31.9 on LVIS), keeping the accuracy on the original datasets (e.g., 65.6 vs 65.5 on COCO, 49.9 vs. 50.2 on LVIS). With RPN proposals, our method also improves on the typographic attack datasets (e.g., 14.4 vs. 11.0 on COCO, 6.25 vs. 5.17 on LVIS) without losing the original performance (e.g., 29.6 vs. 29.6 on COCO, 11.3 vs. 11.1 on LVIS).

\subsection{Ablation Studies}
\paragraph{Effectiveness of our identity loss:}
Table~\ref{table:ablation-origin} lists the effects of the \textit{identity loss}. We observe that the performance of DP trained without identity loss drops drastically on the original datasets (e.g., from 60.91\% to 55.43\% on average). Identity loss effectively helps the learned token maintain the original meanings of the words. Although categorical knowledge distillation has not been commonly used in VLP, the distillation works effectively as a regularization term. 

\begin{table*}[t]
\begin{center}
\small
\caption{\textbf{Ablation studies on the effect of identity loss on original datasets}}
\label{table:ablation-origin}
\begin{tabular}{@{}llllllllllll@{}} \toprule
    Method & ImageNet & Caltech & Pets & Cars & Flowers & Food & Aircraft & DTD & SUN & SAT & Avg. \\ \midrule
    CLIP & 62.02& 88.64& 87.35& 58.72& 66.32& 84.14& 18.99& 44.57& 61.74& 42.98& 61.55\\ \midrule
    Ours w/o identity loss & 55.81& 85.01& 86.67& 52.77& 58.79& 77.89& 15.48& 30.8& 52.2& 38.86& 55.43 \\
    Ours w/ identity loss & \textbf{62.48}& \textbf{89.28}& \textbf{87.22}& \textbf{57.47}& \textbf{63.82}& \textbf{83.65}& \textbf{19.26}& \textbf{40.64}& \textbf{61.41}& \textbf{43.85}& \textbf{60.91} \\ \bottomrule
\end{tabular}
\end{center}
\end{table*}

\begin{table}[t]
\begin{center}
% \small
\caption{\textbf{Ablation studies on the position of the DP token}}
\label{table:ablation-ta-position}
\begin{tabular}{@{}llll@{}} \toprule
     & & \multicolumn{2}{c}{Typographic attack} \\ \cmidrule(l) {3-4} 
    The position & Original & Synth. & Real \\ \midrule
    the beginning & 60.50 & 44.13 & 63.11  \\
    the end & \textbf{61.09} & 37.82 & 55.69 \\
    before class names & 60.91 & \textbf{44.20} & \textbf{64.52} \\ \bottomrule
\end{tabular}
\end{center}
% \end{table*}
% \begin{table}
\begin{center}

% \small
\caption{\textbf{Ablation studies on the number of DP tokens}}
\label{table:ablation-number}
\begin{tabular}{@{}llll@{}} \toprule
     & & \multicolumn{2}{c}{Typographic attack} \\ \cmidrule(l) {3-4} 
    Number of tokens & Original & Synth. & Real \\ \midrule
    one token & \textbf{60.91} & \textbf{44.20} & \textbf{64.52}\\
    two tokens & 59.57 & 43.41 & 60.41 \\
    three tokens & 47.3 & 34.23 & 48.07 \\ \bottomrule
\end{tabular}
\end{center}
\end{table}

\paragraph{Position of the DP token:}

There are many possible positions for the placement of the DP token. These include: at the beginning of a sentence~\cite{kuniaki2020prefix}, before a class name~\cite{Ruiz2022-oy, kumari2022customdiffusion}, and at the end of a sentence.

Table~\ref{table:ablation-ta-position} shows the effect of the position of DP. We observe that the performance of DP at the beginning and end of the sentence decreases on synthetic and real-world typographic attack datasets. The result indicates that DP works most effectively before a class name.

% \vspace{-3mm}
\paragraph{The number of DP tokens:}
Table ~\ref{table:ablation-number} shows the effect of the number of DP tokens. When we increase the number of DP tokens, the overall classification accuracy drops. The result indicates that the best number of tokens is one for our DP.

% \vspace{-3mm}
\paragraph{Hyperparameters:}
In Sec.~\ref{defense-prefix}, we use hyperparameters $\lambda$. About the value of $\lambda$, we conduct an ablation study. As Table~\ref{table:hyperparameter} shows, there is no optimal $\lambda$, and we used $\lambda=3.0$. Also, when we train defense-prefix with only identity loss, the performance is similar to original CLIP's score.

\begin{table}[h]
\begin{center}
\caption{\textbf{Ablation study about hyper-parameters}}
\label{table:hyperparameter}
\begin{tabular}{@{}llll@{}} \toprule
    Method & Original & Synth. & Real \\ \midrule
    CLIP & 61.55 & 34.59 & 46.82 \\ \midrule
    w/o defense loss & 61.72 & 35.19 & 51.16\\ 
    $\lambda=2.0$ & 60.93 & \textbf{45.31} & 63.21\\
    $\lambda=2.5$ & \textbf{61.75} & 44.73 & 62.73\\
    $\lambda=3.0$ & 60.91 & 44.20 & 64.52\\
    $\lambda=3.5$ & 61.21 & 44.72 & 64.16\\
    $\lambda=4.0$ & 61.37 & 44.82 & \textbf{64.71} \\\bottomrule
\end{tabular}
\end{center}
\end{table}

\section{Conclusion}
In this study, we tackled reducing the impact of typographic attacks on CLIP. To achieve this, we proposed Defense-Prefix, a novel method for preventing typographic attacks on CLIP. We explored the application of \textit{class-prefix learning}, which is primarily conducted in subject-driven image generation.
To maintain the generalization ability of CLIP, we used categorical knowledge distillation as a regularization loss. This helped the learned prefix maintain the original meanings of the words. Although our method did not require updating CLIP, it effectively prevented typographic attacks on CLIP, while keeping the model's original performance. In addition, we demonstrated that our approach could be easily applied to downstream tasks such as object detection. This is a significant advantage over the existing studies, which require a modification of the model.
% \vspace{-1mm}
\subsection*{Future work \& limitation}
% \vspace{-1mm}
Our method loses to the previous study on synthetic typographic attack datasets. In addition, we only addressed the problem of typographic attacks. We believe that the proposed method can be applied to other adversarial attacks on VLP. We hope that this work will shed light on research on the utilization of VLP.

\newpage
{\small
\bibliographystyle{ieee_fullname}
\bibliography{egbib}
}

\newpage
\onecolumn
\newcommand\beginsupplement{%
        \setcounter{table}{0}
        \renewcommand{\thetable}{\Alph{table}}%
        \setcounter{figure}{0}
        \renewcommand{\thefigure}{\Alph{figure}}%
     }
\beginsupplement
\appendix

\section{Prompts}

In Sec. 3.2, we use templates to prepare input text $t_i$ and $t_i^{\mathrm{DP}}$. For training, we randomly choose a template from hand-crafted prompts in each iteration. For hand-crafted, we use 81 prompts: (
        `\verb|<CLS>|.',
        `a photo of a \verb|<CLS>|.',
        `a bad photo of a \verb|<CLS>|.',
        `a photo of many \verb|<CLS>|.',
        `a sculpture of a \verb|<CLS>|.',
        `a photo of the hard to see \verb|<CLS>|.',
        `a low resolution photo of the \verb|<CLS>|.',
        `a rendering of a \verb|<CLS>|.',
        `graffiti of a \verb|<CLS>|.',
        `a bad photo of the \verb|<CLS>|.',
        `a cropped photo of the \verb|<CLS>|.',
        `a tattoo of a \verb|<CLS>|.',
        `the embroidered \verb|<CLS>|.',
        `a photo of a hard to see \verb|<CLS>|.',
        `a bright photo of a \verb|<CLS>|.',
        `a photo of a clean \verb|<CLS>|.',
        `a photo of a dirty \verb|<CLS>|.',
        `a dark photo of the \verb|<CLS>|.',
        `a drawing of a \verb|<CLS>|.',
        `a photo of my \verb|<CLS>|.',
        `the plastic \verb|<CLS>|.',
        `a photo of the cool \verb|<CLS>|.',
        `a close-up photo of a \verb|<CLS>|.',
        `a black and white photo of the \verb|<CLS>|.',
        `a painting of the \verb|<CLS>|.',
        `a painting of a \verb|<CLS>|.',
        `a pixelated photo of the \verb|<CLS>|.',
        `a sculpture of the \verb|<CLS>|.',
        `a bright photo of the \verb|<CLS>|.',
        `a cropped photo of a \verb|<CLS>|.',
        `a plastic \verb|<CLS>|.',
        `a photo of the dirty \verb|<CLS>|.',
        `a jpeg corrupted photo of a \verb|<CLS>|.',
        `a blurry photo of the \verb|<CLS>|.',
        `a photo of the \verb|<CLS>|.',
        `a good photo of the \verb|<CLS>|.',
        `a rendering of the \verb|<CLS>|.',
        `a \verb|<CLS>| in a video game.',
        `a photo of one \verb|<CLS>|.',
        `a doodle of a \verb|<CLS>|.',
        `a close-up photo of the \verb|<CLS>|.',
        `the origami \verb|<CLS>|.',
        `the \verb|<CLS>| in a video game.',
        `a sketch of a \verb|<CLS>|.',
        `a doodle of the \verb|<CLS>|.',
        `a origami \verb|<CLS>|.',
        `a low resolution photo of a \verb|<CLS>|.',
        `the toy \verb|<CLS>|.',
        `a rendition of the \verb|<CLS>|.',
        `a photo of the clean \verb|<CLS>|.',
        `a photo of a large \verb|<CLS>|.',
        `a rendition of a \verb|<CLS>|.',
        `a photo of a nice \verb|<CLS>|.',
        `a photo of a weird \verb|<CLS>|.',
        `a blurry photo of a \verb|<CLS>|.',
        `a cartoon \verb|<CLS>|.',
        `art of a \verb|<CLS>|.',
        `a sketch of the \verb|<CLS>|.',
        `a embroidered \verb|<CLS>|.',
        `a pixelated photo of a \verb|<CLS>|.',
        `itap of the \verb|<CLS>|.',
        `a jpeg corrupted photo of the \verb|<CLS>|.',
        `a good photo of a \verb|<CLS>|.',
        `a plushie \verb|<CLS>|.',
        `a photo of the nice \verb|<CLS>|.',
        `a photo of the small \verb|<CLS>|.',
        `a photo of the weird \verb|<CLS>|.',
        `the cartoon \verb|<CLS>|.',
        `art of the \verb|<CLS>|.',
        `a drawing of the \verb|<CLS>|.',
        `a photo of the large \verb|<CLS>|.',
        `a black and white photo of a \verb|<CLS>|.',
        `the plushie \verb|<CLS>|.',
        `a dark photo of a \verb|<CLS>|.',
        `itap of a \verb|<CLS>|.',
        `graffiti of the \verb|<CLS>|.',
        `a toy \verb|<CLS>|.',
        `itap of my \verb|<CLS>|.',
        `a photo of a cool \verb|<CLS>|.',
        `a photo of a small \verb|<CLS>|.',
        `a tattoo of the \verb|<CLS>|.',
    )

In Sec. 4.2, we evaluate our method through classification. For classification, we use hand-crafted prompts (Table ~\ref{table:prompt}).

\begin{table*}[t]
\begin{center}
\small
\caption{\textbf{Prompts used for inference}}
\label{table:prompt}
\begin{tabular}{@{}ll@{}} \toprule
    Dataset & Prompt \\ \midrule
    ImageNet & ``a photo of a \verb|<CLS>|.'' \\ 
    Caltech101 & ``a photo of a \verb|<CLS>|.'' \\ 
    OxfordPets & ``a photo of a \verb|<CLS>|, a type of pet.'' \\ 
    StanfordCars & ``a photo of a \verb|<CLS>|.'' \\ 
    Flowers102 & ``a photo of a \verb|<CLS>|, a type of flower.'' \\ 
    Food101 & ``a photo of a \verb|<CLS>|, a type of food.'' \\
    FGVCAircraft & ``a photo of a \verb|<CLS>|, a type of aircraft.'' \\ 
    DTD & ``\verb|<CLS>| texture.'' \\
    SUN397 & ``a photo of a \verb|<CLS>|.'' \\ 
    EuroSAT & ``a centered satellite photo of a \verb|<CLS>|.'' \\ 
    Real-world typographic attack datasets & ``a photo of a \verb|<CLS>|.'' \\ \bottomrule
\end{tabular}
\end{center}
\end{table*}
% \vspace{0.5cm}
% \begin{tabular}{|c|c|} \hline
%     \rowcolor[gray]{0.8}%
%     Dataset & Prompt \\ \hline
%     ImageNet & ``a photo of a \verb|<CLS>|.'' \\ \hline
%     Caltech101 & ``a photo of a \verb|<CLS>|.'' \\ \hline
%     OxfordPets & ``a photo of a \verb|<CLS>|, a type of pet.'' \\ \hline
%     StanfordCars & ``a photo of a \verb|<CLS>|.'' \\ \hline
%     Flowers102 & ``a photo of a \verb|<CLS>|, a type of flower.'' \\ \hline
%     Food101 & ``a photo of a \verb|<CLS>|, a type of food.'' \\ \hline
%     FGVCAircraft & ``a photo of a \verb|<CLS>|, a type of aircraft.'' \\ \hline
%     DTD & ``\verb|<CLS>| texture.'' \\ \hline
%     SUN397 & ``a photo of a \verb|<CLS>|.'' \\ \hline
%     EuroSAT & ``a centered satellite photo of a \verb|<CLS>|.'' \\ \hline
%     Real-world typographic attack datasets & ``a photo of a \verb|<CLS>|.'' \\ \hline
% \end{tabular}
% \vspace{0.5cm}

\section{Synthetic typographic attack datasets}
    \label{appendix:datasets}

\begin{figure}[t]
\begin{center}
   \includegraphics[width=1.0\linewidth]{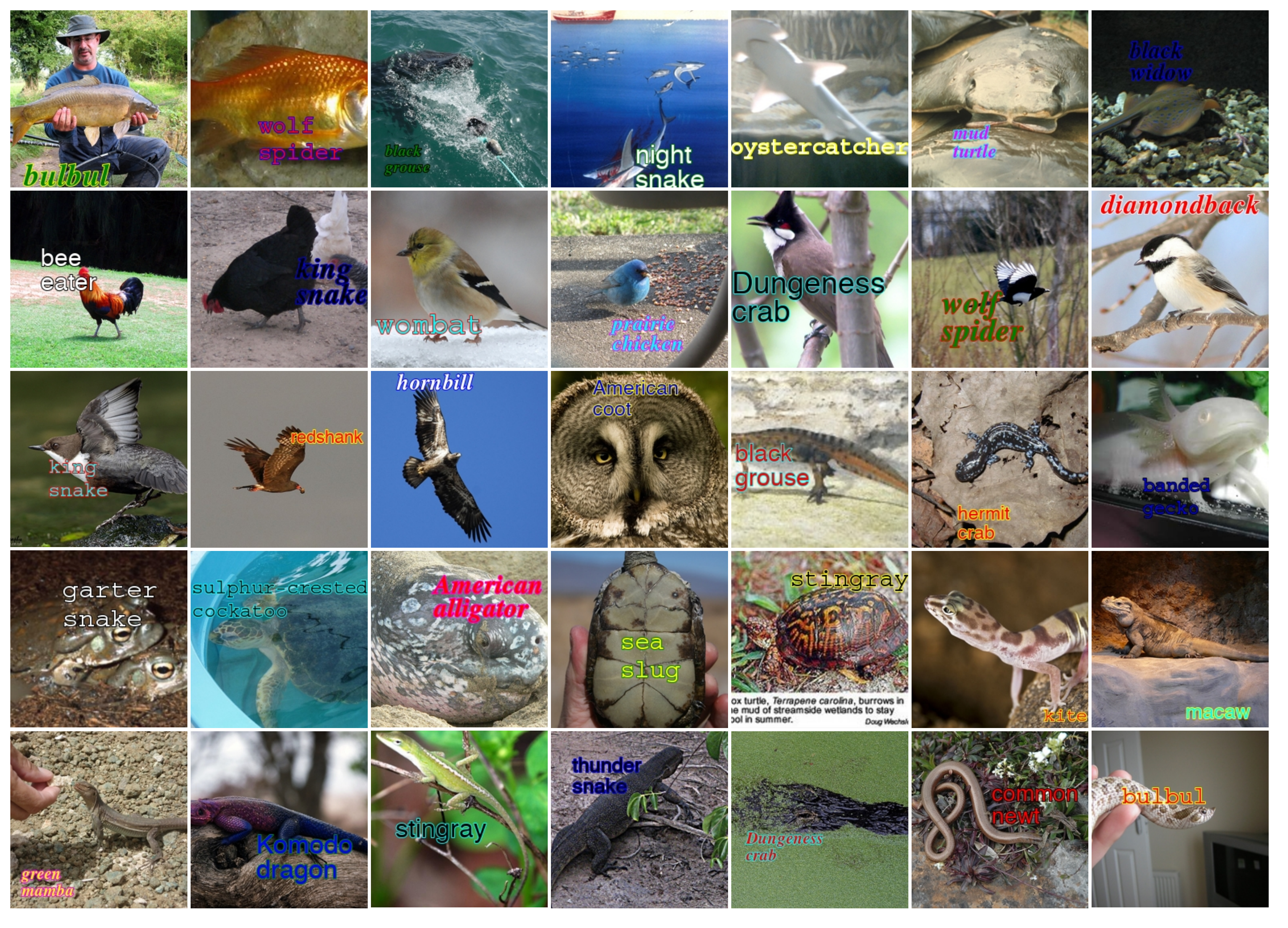}
\end{center}
   \caption{\textbf{Images sampled from our training dataset.} The dataset consists of images from ImageNet-100 with synthesized text.}
\label{fig:train_data}
\end{figure}
\begin{figure}
\begin{center}
   \includegraphics[width=1.0\linewidth]{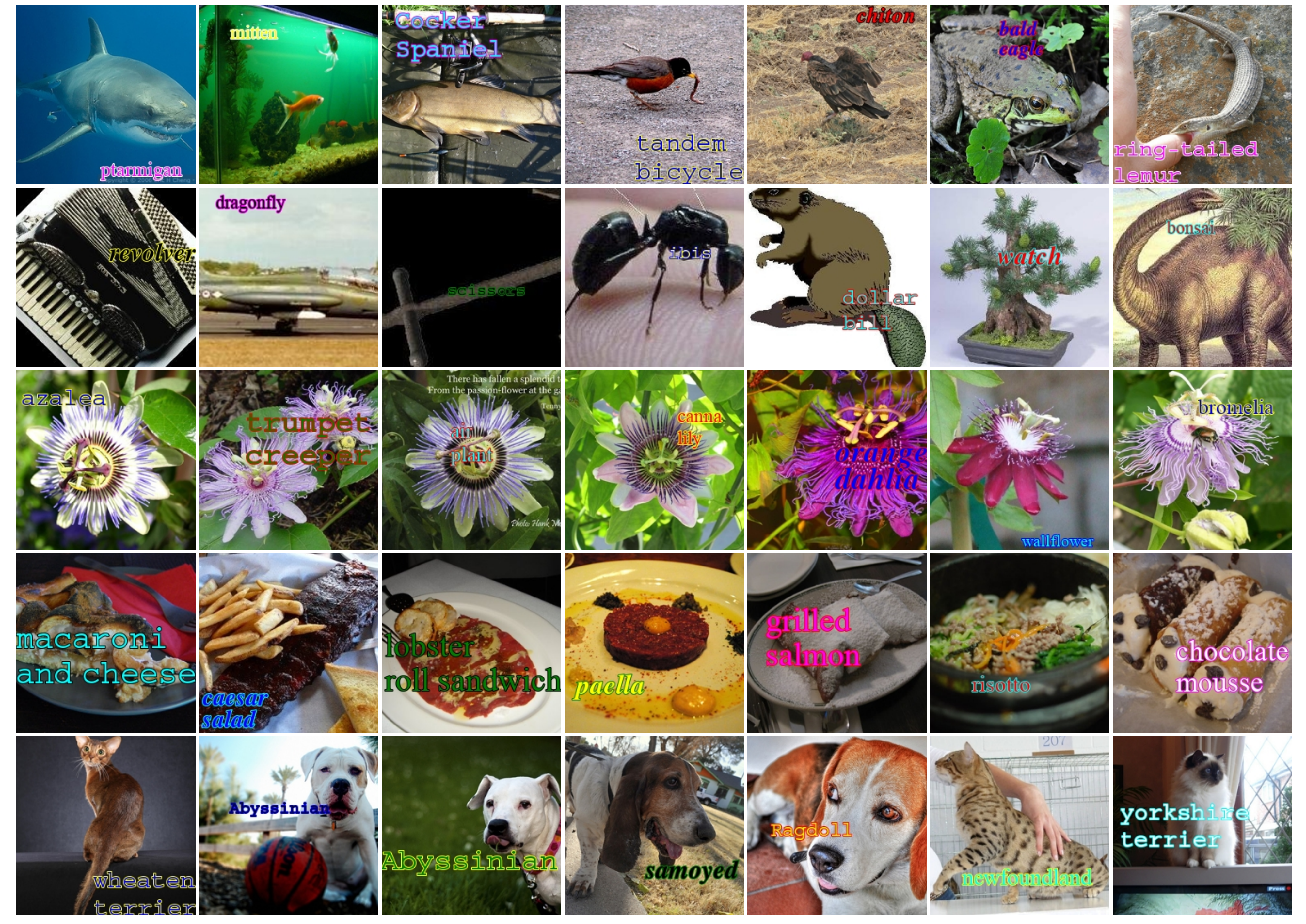}
\end{center}
   \caption{\textbf{Images sampled from our test datasets.} We use ten datasets to make test data for synthetic typographic attack datasets.}
\label{fig:test_data}
\end{figure}

\begin{figure}[t]
\begin{center}
   \includegraphics[width=1.0\linewidth]{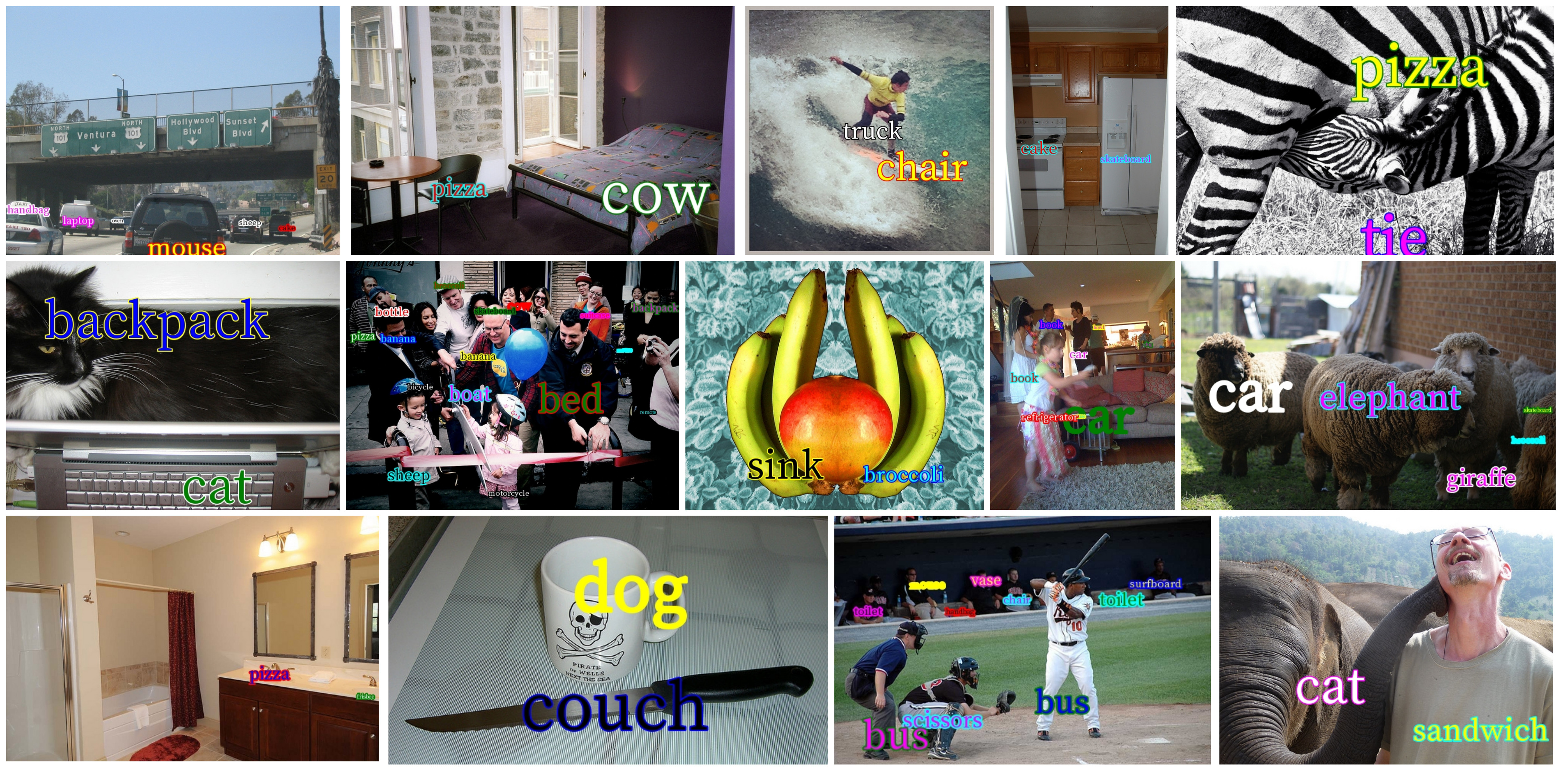}
\end{center}
   \caption{\textbf{Images sampled from our typographic attack COCO dataset.} The dataset consists of images from COCO with synthesized text.}
\label{fig:regionclip_data}
\end{figure}
In this Sec., we will explain the details of the training data in Sec. 3.2 and test data in Sec. 4.2. When we train the DP vector (Sec. 3.2) and conduct experiments on classification (Sec. 4.2), we use synthetic typographic attack datasets. For training data, we add text to images from ImageNet-100 (Figure ~\ref{fig:train_data}). For test data, we add text to images from ten classification datasets (Figure ~\ref{fig:test_data}): ImageNet~\cite{Deng2009-rk}, Caltech101~\cite{Fei2004caltech}, OxfordPets~\cite{Parkhi2012pets}, StanfordCars~\cite{Krause2013cars}, Flowers102~\cite{Nilsback2008flowers}, Food101~\cite{Bossard2014food}, FGVCAircraft~\cite{Maji2013aircraft}, DTD~\cite{Cimpoi2014dtd}, SUN397~\cite{xial2018sun}, EuroSAT~\cite{helber2019eurosat}.
% For the synthetic typographic attack datasets, we add text to images. 
 To make typographic attack datasets, we followed the way of PAINT~\cite{Ilharco2022}. We resize the short dimension to 224 pixels using bicubic interpolation and crop 224 pixels by 224 pixels in the center, which is the standard CLIP~\cite{Radford2021-bi} resize and crop augmentation. For fonts, we randomly choose from three fonts: Roman, Courier, Times. For font size, we randomly sample between 20 and 40 points. Also, we randomize over eight colors: red, green, blue, cyan, magenta, yellow, white, and black. We outline text with a 1-point shadow that is a different color from the main font color. The text is randomly placed in the image such that whole words are visible. Text is chosen from the class labels of the dataset except for the correct image labels.

For object detection, we also make synthetic typographic attack datasets using COCO~\cite{lin2014coco} and LVIS~\cite{gupta2019lvis} (Figure \ref{fig:regionclip_data}). We use AdobeVFPrototype as a font. We randomize over eight colors: red, green, blue, cyan, magenta, yellow, white, and black. We outline text with a 1-point shadow that is a different color from the main font color. The text is randomly placed in each bounding box such that the whole words are visible. We adjust the font size so that the width of the text is less than 0.8 times the width of the bounding box.

% \vspace{30mm}-
\section{RTA-100}
    \label{appendix:our-dataset}
\begin{figure}[t]
\begin{center}
   \includegraphics[width=1.0\linewidth]{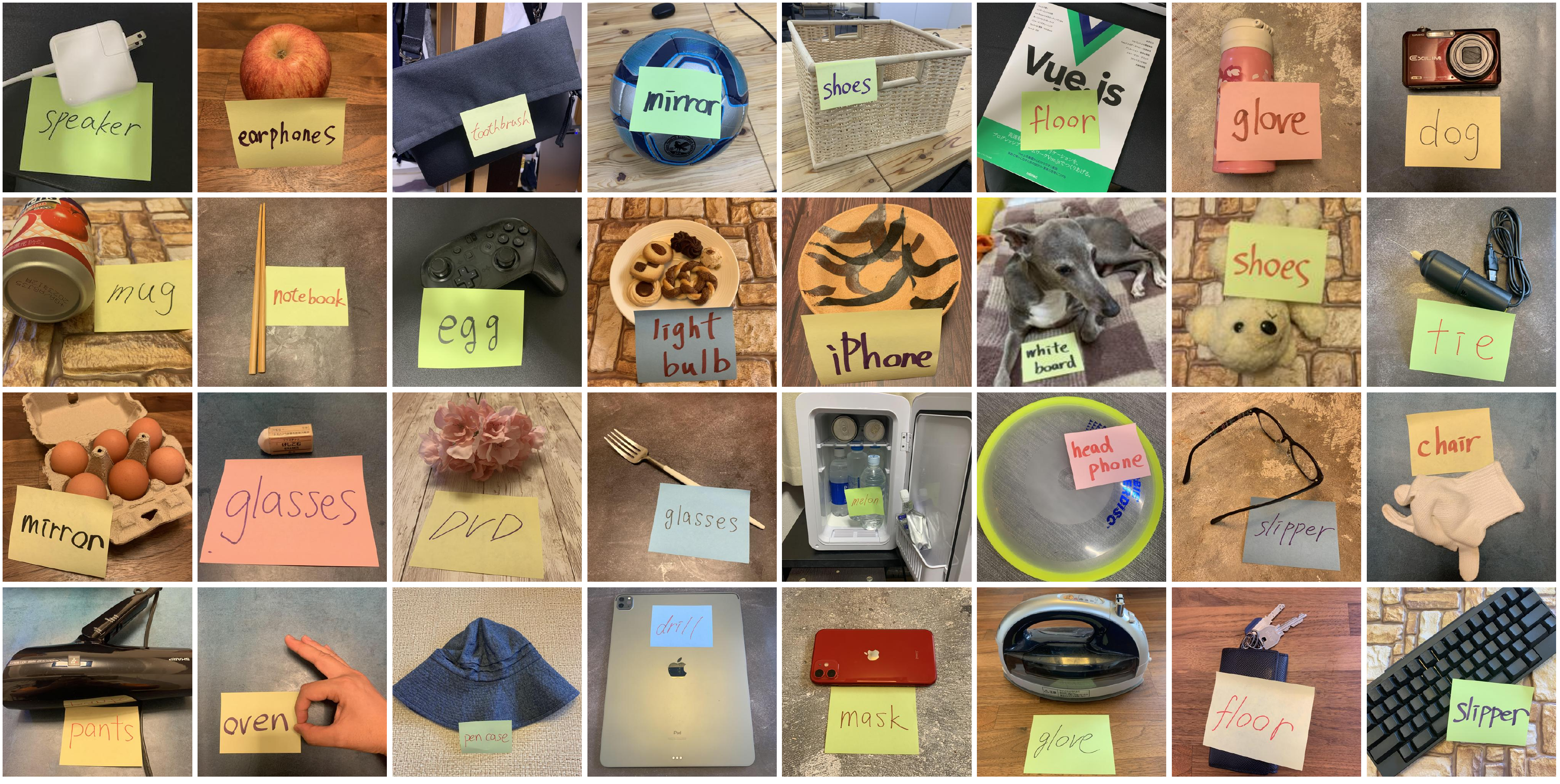}
\end{center}
   \caption{\textbf{Sample images from our real-world typographic attack dataset RTA-100.} The dataset contains 1000 images composed of 100 categories.}
\label{fig:our_data}
\end{figure}
% \chapter{Solution of discretized optimization problems}
% \label{app:SolutionDiscretizedOptProb}
\begin{figure}
\begin{center}
   \includegraphics[width=0.5\linewidth]{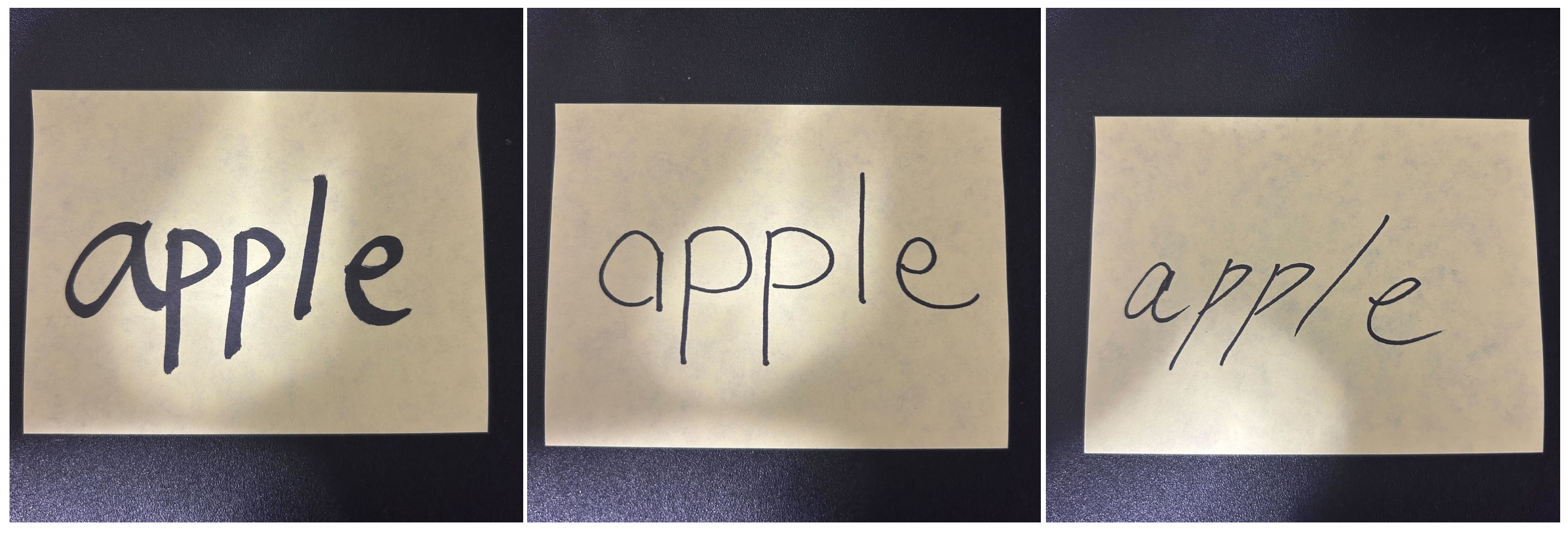}
\end{center}
   \caption{\textbf{Sample images of fonts we used.} We use three fonts to write text: bold, normal, and italic.}
\label{fig:fonts}
\end{figure}

In Sec. 4.2, we use real-world typographic attack datasets. To increase the test data, we take pictures and make RTA-100, which is the biggest real-world typographic attack dataset (Figure ~\ref{fig:our_data}). We put tags that are labeled incorrect classes to objects. We choose the incorrect labels of the tags from the objects in our dataset. We take pictures from 10cm to 2m from the objects such that whole words are visible. For example, we write ``pen'' on the tag and put it on a frisbee. Then, we take a photo of the object. For fonts, we randomly choose from three fonts, as seen in Figure ~\ref{fig:fonts}. For the color of the tags, we randomly choose from 4 colors: yellow, green, blue, and pink. Also, we randomize over 4 colors for the color of the pen: black, red, purple, and brown. We randomly choose these elements in advance. The dataset contains 100 categories and 1000 images. We use iPhoneX's camera, and the size of images is 3024 pixels by 3024 pixels. The code and dataset will be publicly available.

\section{PAINT}
In Sec. 4.2, we compare our method with PAINT~\cite{Ilharco2022}. For training for PAINT, we train the model for 3 epochs (2400 iterations) with batch size 16 using learning rate 1e-5 with 200 warm-up steps with a cosine annealing learning rate schedule and the AdamW optimizer (weight decay 0.1), following the paper.

\section{Extended results on all datasets.}

\begin{table*}[t]
    \begin{center}
\small
\caption{\textbf{Classification results on all original datasets}}
\label{table:original}
\begin{tabular}{@{}lllllllllllll@{}} \toprule
    Method & ImageNet & Caltech & Pets & Cars & Flowers & Food & Aircraft & DTD & SUN & SAT & Avg. \\ \midrule
    CLIP & 62.02& 88.64& 87.35& 58.72& 66.32& 84.14& 18.99& 44.57& 61.74& 42.98& 61.55 \\ \midrule
    Materzynska+~\cite{Materzynska2022-pf} & 54.38& 80.53& 75.01& 40.33& 51.86& 55.01& 13.23& 36.28& 51.06& 37.32& 49.50 \\
    PAINT~\cite{Ilharco2022} & 61.82& 88.48& 85.23& 55.30& \textbf{64.73}& 80.51& 17.73& \textbf{42.61}& \textbf{61.69}& 38.20& 59.63 \\
    Ours & \textbf{62.48}& \textbf{89.28}& \textbf{87.22}& \textbf{57.47}& 63.82& \textbf{83.65}& \textbf{19.26}& 40.64& 61.41& \textbf{43.85}& \textbf{60.91} \\ \bottomrule
\end{tabular}
\end{center}
% \end{table*}

% \begin{table*}[t]
\begin{center}
\vspace{-1mm}
\small
\caption{\textbf{Classification results on all synthetic typographic attack datasets}}
\label{table:ta}
\begin{tabular}{@{}llllllllllllll@{}} \toprule
    Method & ImageNet & Caltech & Pets & Cars & Flowers & Food & Aircraft & DTD & SUN & SAT & Avg. \\ \midrule
    CLIP & 39.10& 63.97& 58.95& 21.02& 31.32& 56.27& 10.83& 25.53& 34.02& 4.86& 34.59\\ \midrule
    Materzynska+~\cite{Materzynska2022-pf} & 44.91& 74.73& 63.61& 15.79& 34.95& 43.41& 8.28& 33.03& 39.52& 16.22& 37.44 \\
    PAINT~\cite{Ilharco2022} & \textbf{55.9}& \textbf{83.57}& \textbf{76.53}& \textbf{33.44}& \textbf{54.92}& \textbf{72.94}& 14.46& \textbf{36.60}& \textbf{53.62}& \textbf{17.31}& \textbf{49.93} \\
    Ours & 49.83& 79.54& 72.88& 28.64& 44.12& 67.79& \textbf{14.49}& 31.6& 43.50& 9.65& 44.20 \\ \bottomrule
\end{tabular}
\end{center}
\end{table*}

In tables ~\ref{table:original} and \ref{table:ta}, we report the accuracy obtained on each of the 10 individual datasets for original and synthetic typographic attacks respectively.

\section{Visualization}
\begin{figure*}[t]
% \vspace{-5mm}
\begin{center}
    \includegraphics[width=0.8\linewidth]{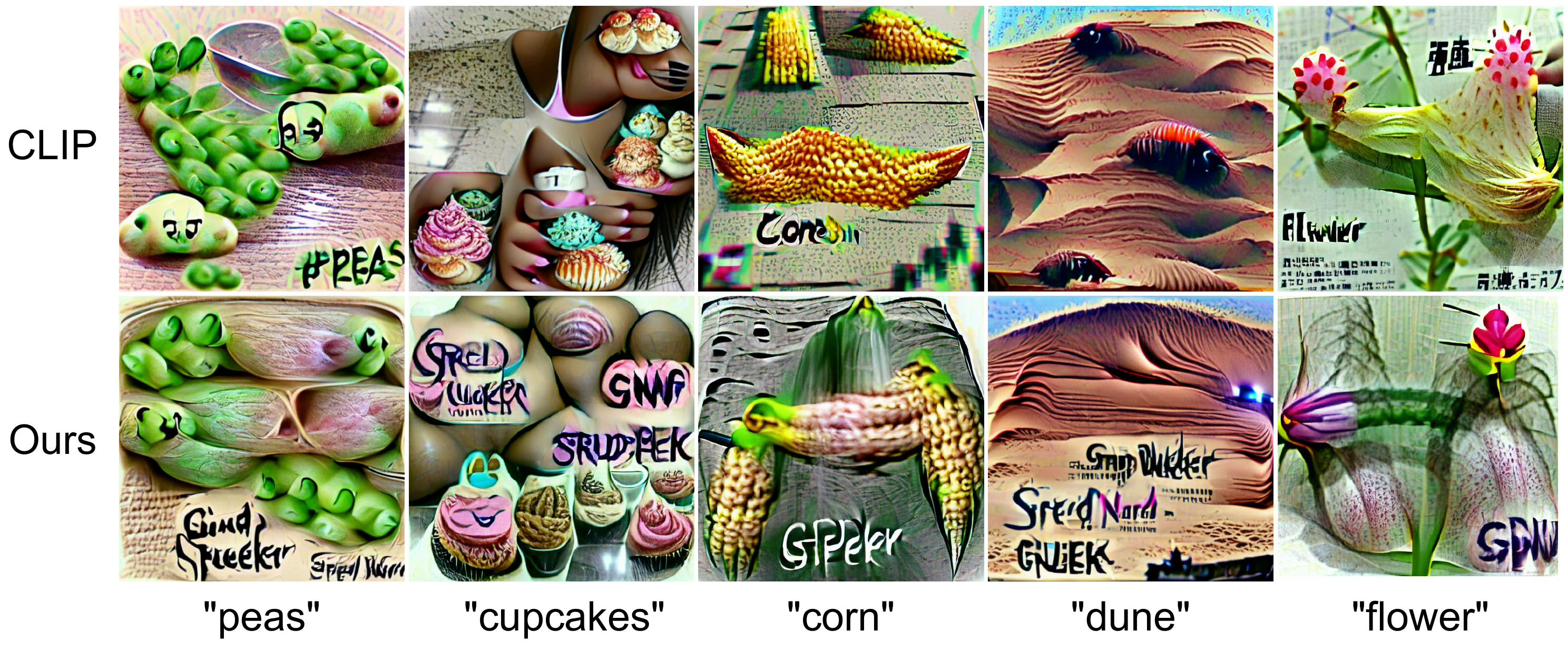}
\end{center}
    % \vspace{-3mm}
    \caption{\textbf{Generated images conditioned on text prompts using VQGAN+CLIP.} Originally, CLIP often generates text of prompts as it is (top row) (e.g., ``peas'', ``corn'', ``flower''). CLIP+Ours does not generate prompt texts in images, showing nonsense strings (bottom row).}
\label{fig:vqgan}
\end{figure*}

To visualize the changes in word information, we generate images conditioned on text prompts using VQGAN+CLIP~\cite{crowson2022vqgan}. Fig.~\ref{fig:vqgan} presents samples of generated images: the first row shows images generated with original VQGAN+CLIP, capturing the visual concepts of the prompt texts. In cases of ``peas'', ``corn'', and ``flower'', the images show the words of the prompts. The images generated with VQGAN+CLIP+Ours can also capture the visual concepts and do not show prompt text; instead, they show nonsense strings. The experiment demonstrates words with DP lose little original meanings, ruining the text information.

\end{document}